\def\year{2020}\relax
\documentclass[letterpaper]{article} 
\usepackage{aaai20}  
\usepackage{times}  
\usepackage{helvet} 
\usepackage{courier}  
\usepackage[hyphens]{url}  
\usepackage{graphicx} 
\usepackage{algorithm}
\usepackage{microtype} 
\usepackage{algpseudocode}
\usepackage{mathrsfs}
\usepackage{comment}

\usepackage{color}

\def\E{{\rm E}}

\title{Motion-Based Generator Model: Unsupervised Disentanglement of Appearance, Trackable and Intrackable Motions in Dynamic Patterns}
\author{Jianwen Xie $^{1 \ast}$, Ruiqi Gao $^{2 \ast}$, Zilong Zheng $^{2}$, Song-Chun Zhu $^{2}$, Ying Nian Wu $^{2}$\\
$^{1}$Hikvision Research Institute, Santa Clara, USA  \hspace{2mm} $^{2}$University of California, Los Angeles, USA\\
}

\begin{document}
\maketitle

\begin{abstract}

Dynamic patterns are characterized by complex spatial and motion patterns. Understanding dynamic patterns requires a disentangled representational model that separates the factorial components. A commonly used model for dynamic patterns is the state space model, where the state evolves over time according to a transition model and the state generates the observed image frames according to an emission model. To model the motions explicitly, it is natural for the model to be based on the motions or the displacement fields of the pixels.  Thus in the emission model, we let the hidden state generate the displacement field,  which warps the trackable component in the previous image frame to generate the next frame while adding a simultaneously emitted residual image to account for the change that cannot be explained by the deformation. The warping of the previous image is about the trackable part of the change of image frame, while the residual image is about the intrackable part of the image. We use a maximum likelihood algorithm to learn the model parameters that iterates between inferring latent noise vectors that drive the transition model and updating the parameters given the inferred latent vectors. Meanwhile we adopt a regularization term to penalize the norms of the residual images to encourage the model to explain the change of image frames by trackable motion. Unlike existing methods on dynamic patterns, we learn our model in unsupervised setting without ground truth displacement fields or optical flows. In addition, our model defines a notion of intrackability by the separation of warped component and residual component in each image frame. We show that our method can synthesize realistic dynamic pattern, and disentangling appearance, trackable and intrackable motions. The learned  models can be useful for motion transfer, and it is natural to adopt it to define and measure intrackability of a dynamic pattern.

\end{abstract}

\section{Introduction}

Dynamic patterns are spatiotemporal processes that exhibit complex spatial and motion patterns, such as dynamic texture (e.g., falling waters, burning fires), as well as human facial expressions and movements. A fundamental challenge in understanding dynamic patterns is learning disentangled representations to separate the underlying factorial components of the observations without supervision \cite{bengio2013representation,mathieu2016disentangling}.
For example, given a video dataset of human
facial expressions, a disentangled representation can include the face's appearance attributes (such as color, identity, and gender), the trackable motion attributes (such as movements of eyes, lip, and noise), and the intrackable motion attributes (such as illumination change). A disentangled representation of dynamic patterns is useful in manipulable video generation and calculating video statistics. The goal of this paper is not only to provide a representational model for video generation, but more importantly, for video understanding by disentangling 
appearance, trackable and intrackable motions in an unsupervised manner. 

Studying video complexity is key to understanding motion perception, and also useful for designing metrics to characterize the video statistics. Researchers in the field of psychophysics, e.g., \cite{pylyshyn2006dynamics}, have studied the human perception of motion uncertainty, and found that human vision fails to track the objects when the number of moving objects  increases or their motions are too random. In the field of computer vision, \cite{li2007dynamic} proposes the intrackability concept in the context of surveillance tracking. \cite{gong2012intrackability} defines intrackability quantitatively to measure the uncertainty of tracking an image patch using the entropy of posterior probability on velocities. In this paper, we are also interested in providing a new method to define and measure the intrackability of videos, by  disentangling the trackable and intrackable components in the videos, in the context of the proposed model.

A widely used representational model for dynamic patterns is the state space model, where the hidden state evolves through time according to a transition model, and the state generates the image frames according to an emission model. The original dynamic texture model of \cite{doretto2003dynamic} is such a model where the hidden state is a low-dimensional vector, and both the transition model and the emission model are linear. The model can be generalized to non-linear versions where the non-linear mappings in the transition and emission models can be parametrized by neural nets \cite{XieGaoZhengZhuWu2019}. 

In terms of the underlying physical processes and the perception of the dynamic patterns, they are largely about motions, i.e., movements of pixels or constituent elements, and it is desirable to have a model that is based explicitly on the motions. In this paper, we propose such a motion-based model for dynamic patterns. Specifically, in the emission model, we let the hidden state generate the displacement field, which warps the trackable component in the previous image frame to generate the next frame while adding a simultaneously emitted residual image to account for the change that cannot be explained by the deformation. Thus, each image frame is decomposed into a trackable component that is obtained by warping the previous frame and an intrackable component in the form of the simultaneously generated residual image. 

We use the maximum likelihood method to learn the model parameters. The learning algorithm iterates between (1) inferring latent noise vectors that drive the transition model, and (2) updating the parameters given the inferred latent vectors. Meanwhile we adopt a regularization term to penalize the norms of the residual images to encourage the model to explain the change of image frames by motion. Unlike existing methods on dynamic patterns, we learn our model in unsupervised setting without ground truth displacement fields or optical flows. Moreover, with the disentangled representation of a video, we can define a notion of intrackability by comparing  the trackable and intrackable components of the image frames to measure video complexity. 

Experiments show that our method can learn realistic dynamic pattern models, the learned motion can be transferred to testing images with unseen appearances, and intrackability can be quantitatively measured under the proposed representation. 

\textbf{Contribution}. Our contributions are summarized below: (1) We propose a novel representational model of dynamic patterns to disentangle the appearance, trackable and intrackable motions. 
(2) The model can be learned in purely unsupervised setting in that the associated maximum likelihood learning algorithm can learn the model without ground truth or pre-inferred displacement fields or optical flows. (3) The learning algorithm does not rely on an extra assisting network as in VAEs \cite{kingma2013auto,RezendeICML2014,MnihGregor2014} and GANs \cite{goodfellow2014generative}. (4) The experiments show that appearance and motion can be well separated, and the motion can be effectively transferred to a new unseen appearance. (4) With such a representational model, a measure of intrackability can be defined to characterize the video statistics, i.e., video complexity, in the context of the model. 

\section{Related work}

Learning generative models for dynamic textures has been extensively studied in the literature 
\cite{doretto2003dynamic,wang2002generative,wang2004analysis}. For instance, the original model for dynamic texture in \cite{doretto2003dynamic} is a vector auto-regressive model coupled with frame-wise dimension reduction by singular value decomposition. It is linear in both the transition model and the emission model. 

By generalizing the energy-based generative ConvNet model in \cite{XieLuICML}, \cite{xie2017synthesizing} develops an energy-based model where the energy function is parametrized by a spatial-temporal bottom-up ConvNet with multiple layers of non-linear spatial-temporal filters that capture complex spatial-temporal patterns in dynamic textures. The model is learned from scratch by maximizing the log-likelihood of the observed data. \cite{HanTian2018} represents dynamic textures by a top-down spatial-temporal generator model that consists of multiple layers of spatial-temporal kernels. The model is trained via alternative back-propagation algorithm. \cite{xie2016cooperative} proposes a cooperative learning scheme to jointly train the models in \cite{xie2017synthesizing} and \cite{HanTian2018} simultaneously for dynamic texture synthesis. Recently, \cite{XieGaoZhengZhuWu2019} proposes a dynamic generator model that consists of non-linear transition model and non-linear emission model. Unlike the above two models in \cite{xie2017synthesizing} and \cite{HanTian2018}, the model in \cite{XieGaoZhengZhuWu2019} unfolds over time and is a causal model. Our work is based on \cite{XieGaoZhengZhuWu2019} and is an extension of it. Compared to \cite{XieGaoZhengZhuWu2019}, our model in this paper represents dynamic patterns with an unsupervised disentanglement of appearance (pixels), trackable motion (pixel displacement), and intrackable motion (residuals). Therefore our model can animate a static image by directly applying the motion extracted from another video to the static image, even though two appearances are not the same. All models mentioned above can not handle this. Additionally, the intrackable motion provides a new perspective to define and measure the intrackability of videos, which makes our model significantly distinct from and go beyond \cite{XieGaoZhengZhuWu2019}.

 Recently, multiple video generation frameworks based on GANs \cite{goodfellow2014generative} have been proposed. For example, VGAN \cite{vondrick2016generating}, TGAN \cite{saito2017temporal}, and MoCoGAN \cite{tulyakov2018mocogan}.

All of the above methods need to recruit a discriminator with appropriate convolutional architecture to evaluates whether the generated videos are from the training data or the video generator. Our work is not within the domain
of adversarial learning. Unlike GAN-based methods, our model is learned by maximum likelihood without recruiting a discriminator network.

\def\I{{\bf I}}

\section{Model and learning}
\subsection{Motion-based generative model} 
Let $\I = (\I_t, t = 0, 1, ..., T)$ be the observed video sequence of dynamic pattern, where $\I_t$ is a frame at time $t$, and $\I_t$ is defined on the 2D rectangle lattice $D$. The motion-based model for the dynamic patterns consists of the following components: 
%
 \begin{eqnarray}
 && s_{t} = (s^{M}_t, s^{R}_t) = f_1(s_{t-1}, h_t),  \label{eq:t} \\
 && M_t = (\delta(x, y), \forall (x, y) \in D)  = f_2(s^{M}_t), \label{eq:g1} \\
 && R_t = f_3(s^{R}_t), \label{eq:g2} \\
 && I_t = f_4(I_{t-1}, M_t), \label{eq:g3} \\
 && \I_t = I_t + R_t + \epsilon_t, \label{eq:e}
 \end{eqnarray}
where $t = 1, ..., T$. We single out $\I_0$ and discuss it in Equation (\ref{eq:0}) below.

In the above model, $f = (f_i, i = 0, 1, 2, 3)$ are neural networks parameterized by $\theta = (\theta_i, i =0, 1, 2, 3)$. 

Equation (\ref{eq:t}) is the transition model, where $s_t$ is the state vector at time $t$, $h_t \sim {\rm N}(0, I)$ is a hidden Gaussian white noise vector, where $I$ is the identity matrix. $h_t$ are independent over $t$. $f_1$ defines the transition from $s_{t-1}$ to $s_t$. 

The state vector $s_t$ consists of two sub-vectors. One is $s^{M}_t$ for motion. The other is $s^{R}_t$ for residual. While $s^{M}_t$ generates the motion of the trackable part of the image frame $\I_{t-1}$, $S^{R}_t$ generates the non-trackable part of $\I_t$. 

\begin{figure}[t]
\begin{center}
\includegraphics[width=.98\linewidth]{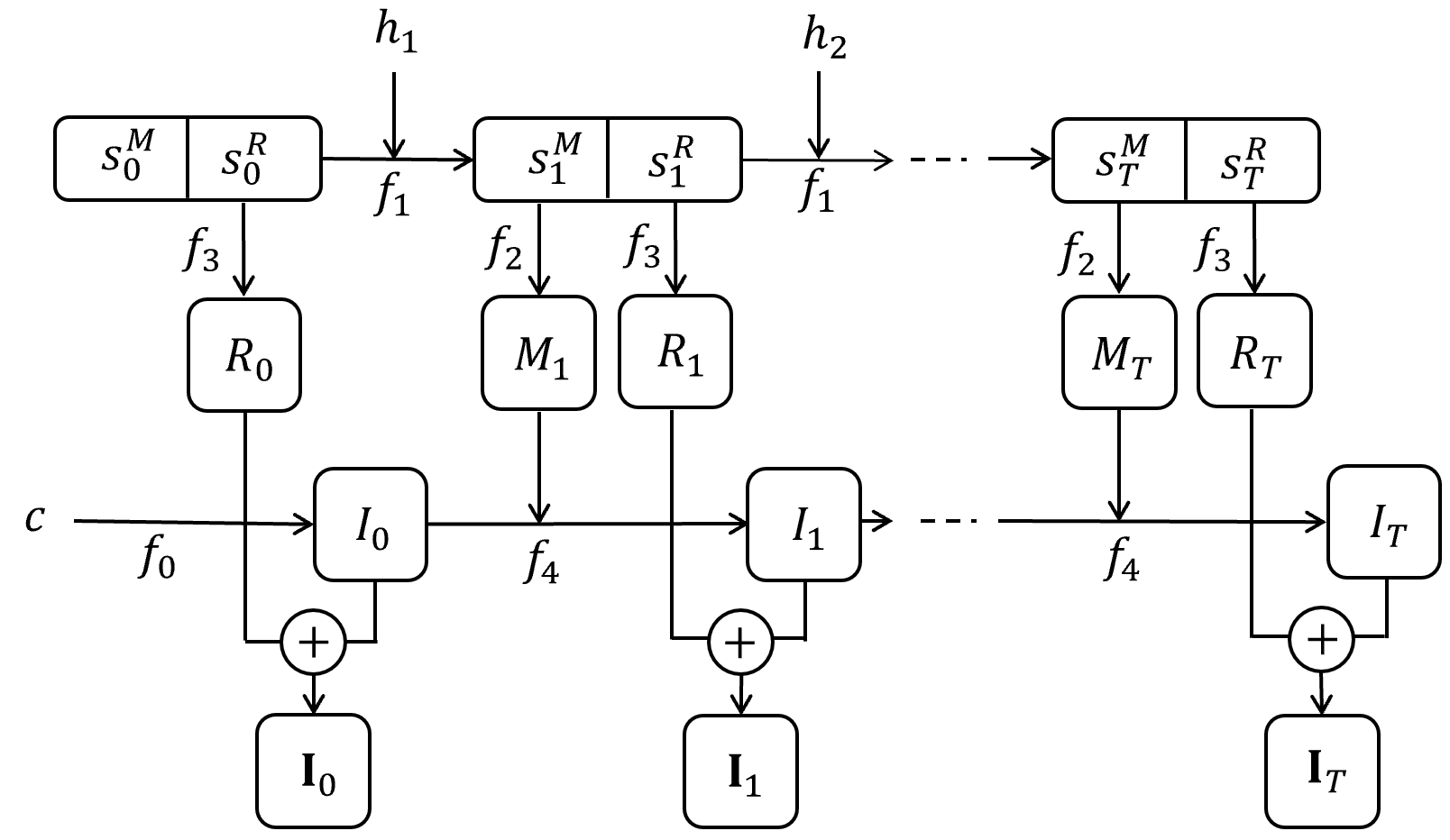}
\caption{An illustration of the framework of the proposed model-based generator model.} 
\label{illustration}
\end{center}
\end{figure}

Specifically, in Equation (\ref{eq:g1}),  $s^{M}_t$ generates the field of pixel displacement $M_t$, which consists of the displacement $\delta(x, y)$ of pixel $(x, y)$ in the image domain $D$. $M_t$ is a 2D image, because the displacement $\delta = (\delta_x, \delta_y)$ is 2D. $f_2$ defines the mapping from $s^{M}_t$ to $M_t$. In Equation (\ref{eq:g3}), $M_t$ is used to warp the trackable part $I_{t-1}$ of the previous image frame $\I_{t-1}$ by a warping function $f_4$, which is given by bilinear interpolation. There is no unknown parameter in $f_4$. In Equation (\ref{eq:g2}), $s^{R}_t$ generates the residual image $R_t$. $f_3$ defines the mapping from $s^{R}_t$ to $R_t$. In Equation (\ref{eq:e}), the image frame $\I_t$ is the sum of the warped image $I_t$ (note that the notation $I_t$ is not in bold font, and it is different from the image frame $\I_t$, which is in bold font) and the residual image $R_t$, plus a Gaussian white noise error $\epsilon_t \sim {\rm N}(0, \sigma^2 I)$. We assume the variance $\sigma^2$ is given. In Equation (\ref{eq:0}), the initial trackable frame $I_0$ is generated by an generator $f_0$ from an appearance hidden variable $c$ that follows Gaussian distribution. 
To initialize the first frame $\I_0$, we use the following method: 
 \begin{eqnarray} 
 I_0 = f_0(c), \; R_0 = f_3(s_0^R), \; \I_0 = I_0 + R_0 + \epsilon_0.
 \label{eq:0} 
 \end{eqnarray}
 Please see Figure \ref{illustration} for an illustration of the proposed model.
 
 {\bf Multiple sequences}. Our model can be easily generalized to handle multiple sequences. We only need to introduce a sequence specific vector $a$, sampled from a Gaussian white noise prior distribution. For each video sequence, this vector $a$ is fixed, and it can be concatenated to the state vector $s_t$ in both the transition model and the emission model. We may also let $a = (a^{M}, a^{R})$, so that $a^{M}$ is concatenated to $s^{M}_t$ to generate $M_t$, and $a^{R}$ is concatenated to $s^{R}_t$ to generate $R_t$. This enables us to disentangle motion pattern and appearance pattern in the video sequence. 

{\bf Intrackability}. For the image $\I_t$, we define $I_t$ to be the trackable part because it is obtained by the movements of pixels, and we define $R_t$ to be the non-trackable part. The intrackability of the sequence can be defined as the ratio between the average of the $\ell_2$ norm of the non-trackable part $R_t$ and the norm of the image $\I_t$, where the average is over the time frames. 

{\bf Summarized form}. Let $h = (h_t, t = 1, ..., T)$. $h$ consist of the hidden random vectors that need to be inferred from $\I = (\I_t, t = 0,1, ..., T)$. We can also include the latent variables $c$ and $s_0$ into $h$ for notation simplicity. Although $\I_t$ is generated by the state vector $s_t$, $s = (s_t, t = 0,1, ..., T)$ are generated by $h$.  In fact, we can write $\I = f_\theta(h) + \epsilon$, where $f_\theta$ composes $f_0$, $f_1$, $f_2$, $f_3$ and $f_4$ over time $t$, and $\epsilon = (\epsilon_t, t = 0, 1, ..., T)$ denotes the observation errors. 

\subsection{Maximum likelihood learning algorithm} 

The model is a generator model with $h$ being the hidden vector. In recent literature, such a model is commonly learned by VAE \cite{kingma2013auto,RezendeICML2014,MnihGregor2014} and GAN \cite{goodfellow2014generative}. However, unlike a regular generator model, $h$ is a sequence of hidden vectors, and we need to design highly sophisticated inference network or discriminator network if we want to implement VAE or GAN, and this is not an easy task.  In this paper, we choose to learn the model by maximum likelihood algorithm which is simple and efficient, without the need to recruit an extra inference or discriminator network. 

Our maximum likelihood learning method is adapted from the recent work \cite{XieGaoZhengZhuWu2019}. Specifically, let $p(h)$ be the Gaussian white noise prior distribution of $h$. Let $p_\theta(\I|h) \sim {\rm N}(f_\theta(h), \sigma^2 I)$ be the conditional distribution of the video sequence $\I$ given $h$. The marginal distribution of $\I$ is $p_\theta(\I) = \int p(h) p_\theta(\I|h) dh$ with the latent variable $h$ integrated out. The log-likelihood is $\log p_\theta(\I)$, which is analytically intractable due to the integral over $h$.  The gradient of the log-likelihood can be computed using the following identity: 
\begin{eqnarray} 
   \frac{\partial}{\partial \theta} \log p_\theta(\I)&=&\frac{1}{p_\theta(\I)} \frac{\partial}{\partial \theta} p_\theta(\I) \nonumber\\ 
   &=& \int \left[ \frac{\partial}{\partial \theta} \log p_\theta(h, \I) \right]  p_\theta(h | \I) dh \nonumber\\
   &=& \E_{p_\theta(h|\I)} \left[  \frac{\partial}{\partial \theta} \log p_\theta(h, \I)  \right],  \label{eq:gr}
\end{eqnarray}
where $p_\theta(h|\I) = p_\theta(h, \I)/p_\theta(\I)$ is the posterior distribution of the latent $h$ given the observed $\I$. The expectation with respect $p_\theta(h|\I)$ can be approximated by Monte Carlo sampling. The sampling of $p_\theta(h|\I)$ can be accomplished by the Langevin dynamics: 
\begin{eqnarray} 
    h^{(\tau+1)} = h^{(\tau)}   + \frac{\delta^2}{2} \frac{\partial}{\partial h} \log p_\theta(h^{(\tau)}|\I) + \delta z_\tau, \label{eq:Langevin}
\end{eqnarray}
where $\tau$ indexes the time step of the Langevin dynamics. Here we use the notation $\tau$ because we have used $t$ to index the time of the video sequence. $h^{(\tau)} = (h^{(\tau)}_{t}, t = 1, ..., T)$. $z_\tau \sim {\rm N}(0, I)$ is the Gaussian white noise vector. $\delta$ is the step size of the Langevin dynamics.  After sampling $h \sim p_\theta(h|\I)$ using the Langevin dynamics, we can  update $\theta$ by stochastic gradient ascent 
\begin{eqnarray} 
     \Delta \theta \propto   \frac{\partial}{\partial \theta} \log p_\theta(h, \I),  \label{eq:learn}
\end{eqnarray}
where we use the sampled $h$ to approximate the expectation in  (\ref{eq:gr}). 

The learning algorithm iterates the following two steps. (1) Inference step: Given the current $\theta$, sample $h$ from $p_\theta(h|\I)$ according to (\ref{eq:Langevin}). (2) Learning step: Given $h$, update $\theta$ according to (\ref{eq:learn}). We can use a warm start to sample $h$ in step (1), that is, when running the Langevin dynamics, we start from the current $h$, and run a finite number of steps. Then we update $\theta$ in step (2) using the sampled $h$. Such a stochastic gradient ascent algorithm has been analyzed by \cite{younes1999convergence}. 

Since $\frac{\partial}{\partial h} \log p_\theta(h|\I) = \frac{\partial}{\partial h} \log p_\theta(h, \I)$, both steps (1) and (2) are based on computing the derivatives of 
\begin{eqnarray}
    \log p_\theta(h, \I) = -\frac{1}{2} \left[ \|h\|^2 + \frac{1}{\sigma^2} \|\I - f_\theta(h)\|^2\right] + {\rm const}, \nonumber
 \end{eqnarray}
where the constant term does not depend on $h$ or $\theta$. The derivatives with respect to $h$ and $\theta$ can be computed efficiently and conveniently by back-propagation through time. 

To encourage the model to explain the video sequence $\I$ by the trackable motion, we add to the log-likelihood $\log p_\theta(\I)$ a penalty term  $- \lambda_1\|R_t\|^2$. To encourage the smoothness of the inferred displacement field $M_t$, we also add another penalty term $-\lambda_2 \|\Delta M_t\|^2$. We estimate $\theta$ by gradient ascent on $\log p_\theta(\I) - \lambda_1 \sum_t \|R_t\|^2 - \lambda_2 \sum_t \|\Delta M_t\|^2$. 

In VAE, we need to define an inference model $q_\phi(h|\I)$ to approximate the posterior distribution $p_\theta(h|\I)$. Due to the complex structure of the model, it is not an easy task to design an accurate inference model. While VAE maximizes a lower bound of the log-likelihood $\log p_\theta(\I)$, where the tightness of the lower bound depends on the Kullback-Leibler divergence between $q_\phi(h|\I)$ and $p_\theta(h|\I)$, our learning algorithm seeks to maximize the log-likelihood itself.

\section{Experiments}

Our paper studies learning to disentangle appearance, trackable motion, and intrackable motion of dynamic pattern in an unsupervised manner by proposing a motion-based dynamic generator. We conduct the following three experiments to test and understand the proposed model. As a generative model for videos, Experiment 1 investigates how good the proposed model can be learned by evaluating its data generation capacity, which is a commonly used way to check whether the learned model can capture the target data distribution. Experiment 2 investigates if the proposed model can successfully decompose the appearance and motion by a task of motion transfer. Experiment 3 studies the disentanglement of trackable and intrackable motions, and use the intrackable one to define the concept of “intrackability”, which is an application of our model.

\begin{figure}[t]
\begin{center} \hspace{0.08mm}
\hspace{0.5mm}\rotatebox{90}{\hspace{4mm}{\footnotesize obs1 }}
\includegraphics[width=.14\linewidth]{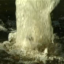}
\includegraphics[width=.14\linewidth]{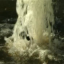}
\includegraphics[width=.14\linewidth]{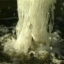}
\includegraphics[width=.14\linewidth]{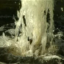}
\includegraphics[width=.14\linewidth]{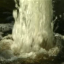}
\includegraphics[width=.14\linewidth]{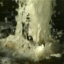}
\\ \vspace{0.5mm}
\hspace{0.5mm}\rotatebox{90}{\hspace{4mm}{\footnotesize syn1 }}
\includegraphics[width=.14\linewidth]{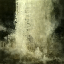}
\includegraphics[width=.14\linewidth]{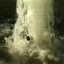}
\includegraphics[width=.14\linewidth]{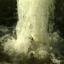}
\includegraphics[width=.14\linewidth]{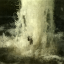}
\includegraphics[width=.14\linewidth]{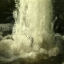}
\includegraphics[width=.14\linewidth]{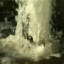}
\\ \vspace{0.5mm}
\hspace{0.5mm}\rotatebox{90}{\hspace{4mm}{\footnotesize syn2 }}
\includegraphics[width=.14\linewidth]{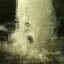}
\includegraphics[width=.14\linewidth]{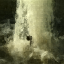}
\includegraphics[width=.14\linewidth]{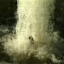}
\includegraphics[width=.14\linewidth]{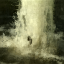}
\includegraphics[width=.14\linewidth]{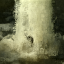}
\includegraphics[width=.14\linewidth]{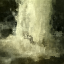}
\\ 
(a) fountain \\ \vspace{0.5mm}
\hspace{0.08mm}
\hspace{0.5mm}\rotatebox{90}{\hspace{4mm}{\footnotesize obs1 }}
\includegraphics[width=.14\linewidth]{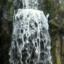}
\includegraphics[width=.14\linewidth]{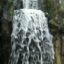}
\includegraphics[width=.14\linewidth]{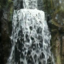}
\includegraphics[width=.14\linewidth]{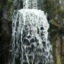}
\includegraphics[width=.14\linewidth]{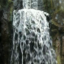}
\includegraphics[width=.14\linewidth]{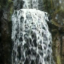}\\ \vspace{0.5mm}
\hspace{0.5mm}\rotatebox{90}{\hspace{4mm}{\footnotesize syn1 }}
\includegraphics[width=.14\linewidth]{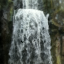}
\includegraphics[width=.14\linewidth]{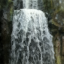}
\includegraphics[width=.14\linewidth]{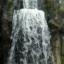}
\includegraphics[width=.14\linewidth]{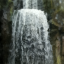}
\includegraphics[width=.14\linewidth]{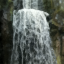}
\includegraphics[width=.14\linewidth]{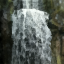}\\ \vspace{0.5mm} \hspace{0.5mm}\rotatebox{90}{\hspace{4mm}{\footnotesize syn2 }}
\includegraphics[width=.14\linewidth]{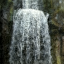}
\includegraphics[width=.14\linewidth]{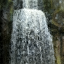}
\includegraphics[width=.14\linewidth]{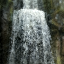}
\includegraphics[width=.14\linewidth]{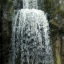}
\includegraphics[width=.14\linewidth]{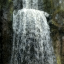}
\includegraphics[width=.14\linewidth]{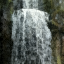}\\ 
(b) waterfall \\ \vspace{0.5mm}
\hspace{0.08mm}
\hspace{0.5mm}\rotatebox{90}{\hspace{4mm}{\footnotesize obs1 }}
\includegraphics[width=.14\linewidth]{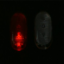}
\includegraphics[width=.14\linewidth]{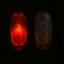}
\includegraphics[width=.14\linewidth]{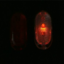}
\includegraphics[width=.14\linewidth]{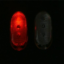}
\includegraphics[width=.14\linewidth]{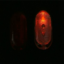}
\includegraphics[width=.14\linewidth]{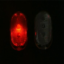}
\\ \vspace{0.5mm}
\hspace{0.5mm}\rotatebox{90}{\hspace{4mm}{\footnotesize syn1 }}
\includegraphics[width=.14\linewidth]{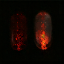}
\includegraphics[width=.14\linewidth]{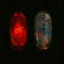}
\includegraphics[width=.14\linewidth]{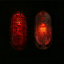}
\includegraphics[width=.14\linewidth]{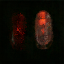}
\includegraphics[width=.14\linewidth]{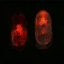}
\includegraphics[width=.14\linewidth]{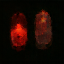}\\
\vspace{0.5mm} \hspace{0.5mm}\rotatebox{90}{\hspace{4mm}{\footnotesize syn2 }}
\includegraphics[width=.14\linewidth]{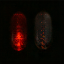}
\includegraphics[width=.14\linewidth]{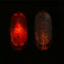}
\includegraphics[width=.14\linewidth]{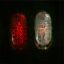}
\includegraphics[width=.14\linewidth]{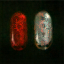}
\includegraphics[width=.14\linewidth]{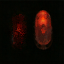}
\includegraphics[width=.14\linewidth]{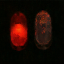}\\
(c) flashing lights 
\caption{Generating dynamic textures. For each category, the first row shows 6 frames of the observed sequence, and the second and third rows show the corresponding frames of two synthesized sequences generated by the learned model. The image size is $64 \times 64$ pixels. The length of each synthesized sequence is 30.}
\label{fig:synthesis}
\end{center}
\end{figure}

\subsection{Implementation details}

Our model was implemented using Python with TensorFlow \cite{tensorflow2015-whitepaper}. Each prepared training video clip is of the size $64 \times 64$ pixels $\times 30$ frames. The configuration of our model architecture is presented as follows. 

\textit{Transition model.} The transition model is a three-layer feedforward neural network that takes a 80-dimensional state vector $s_{t-1}$ and a 100-dimensional noise vector $h_t$ at time $t-1$ as input and outputs a 80-dimensional vector $r_t$ at time $t$, so that $s_{t} = \tanh(s_{t-1} + r_t)$. This is a residual form \cite{he2016deep} for computing $s_{t}$ given $s_{t-1}$. The output of each of the first two layers is followed by a ReLU operation. The tanh activation function is crucial to prevent $s$ from being increasingly lager during the recurrent computation of $s$ by constraining it within the range of $[-1,1]$. The numbers of nodes in the three layers of the feedforward neural network are $\{20, 20, 100\}$. Each state vector consists of two parts $s=[s^{M}, s^{R}]$, where $s^{M}$ is a 50-dimensional motion state vector and $s^{R}$ is a 30-dimensional residual state vector. 

\textit{Emission model}. The emission model for motion is a top-down deconvolution neural network or generator model that maps $s^{M}$ (i.e., $1 \times 1 \times 50$) to the displacement field or optical flow of size $64 \times 64 \times 2$ by 6 layers of deconvolutions with kernel size 4 and up-sampling factor from top to bottom. The numbers of channels at different layers of the generator are $\{512, 512, 256, 128, 64, 2\}$ from top to bottom. Batch normalization \cite{ioffe2015batch} and ReLU layers are added between deconvolution layers, and tanh activation function is used at the bottom layer to make the output signals fall within $[-1, 1]$. The emission model for residuals is also a generator model that maps $s^{R}$ (i.e., $1 \times 1 \times 30$) to the residual image frame of size $64 \times 64 \times 3$ by a generator with the same structure as the emission model for motion, except that the output channel of the last layer is 3 rather than 2. A generator of trackable appearance that maps a 10-dimensional noise vector to the  first image frame follows the same structure as the residual generator.

\textit{Optimization and inference.} Adam \cite{kingma2015adam} is used for optimization with $\beta_1= 0.5$ and the learning rate is 0.001. The Langevin step size is set to be $\delta=0.03$ for all latent variables, and the standard deviation of residual error $\sigma=0.5$. During each learning iteration, we run $15$ steps of Langevin dynamics for inferring the latent noise vectors. Unless otherwise stated, the penalty weights for residuals and smoothness of the displacement field are set to be $\lambda_1=1$ and $\lambda_2=0.005$, respectively.

\subsection{Experiment 1: Dynamic pattern synthesis}

 We firstly evaluate the representational power of the proposed model by applying it to dynamic pattern sythesis. A good generative model for video should be able to generate samples that are perceptually indistinguishable from the real training videos in terms of appearance and dynamics. We learn our models from a wide range of dynamic textures (e.g., flowing water, fire, etc), which are selected from DynTex++ dataset of \cite{ghanem2010maximum} and the Internet. We learn a single model from each training example and generate multiple synthesized examples by simply drawing independent and identically distributed samples from Gaussian distribution of the latent factors. Note that our model only learns from raw video data without relying on other information, such as optical flow ground truths.

\begin{figure}[t!]
\begin{center}
\includegraphics[width=.8\linewidth]{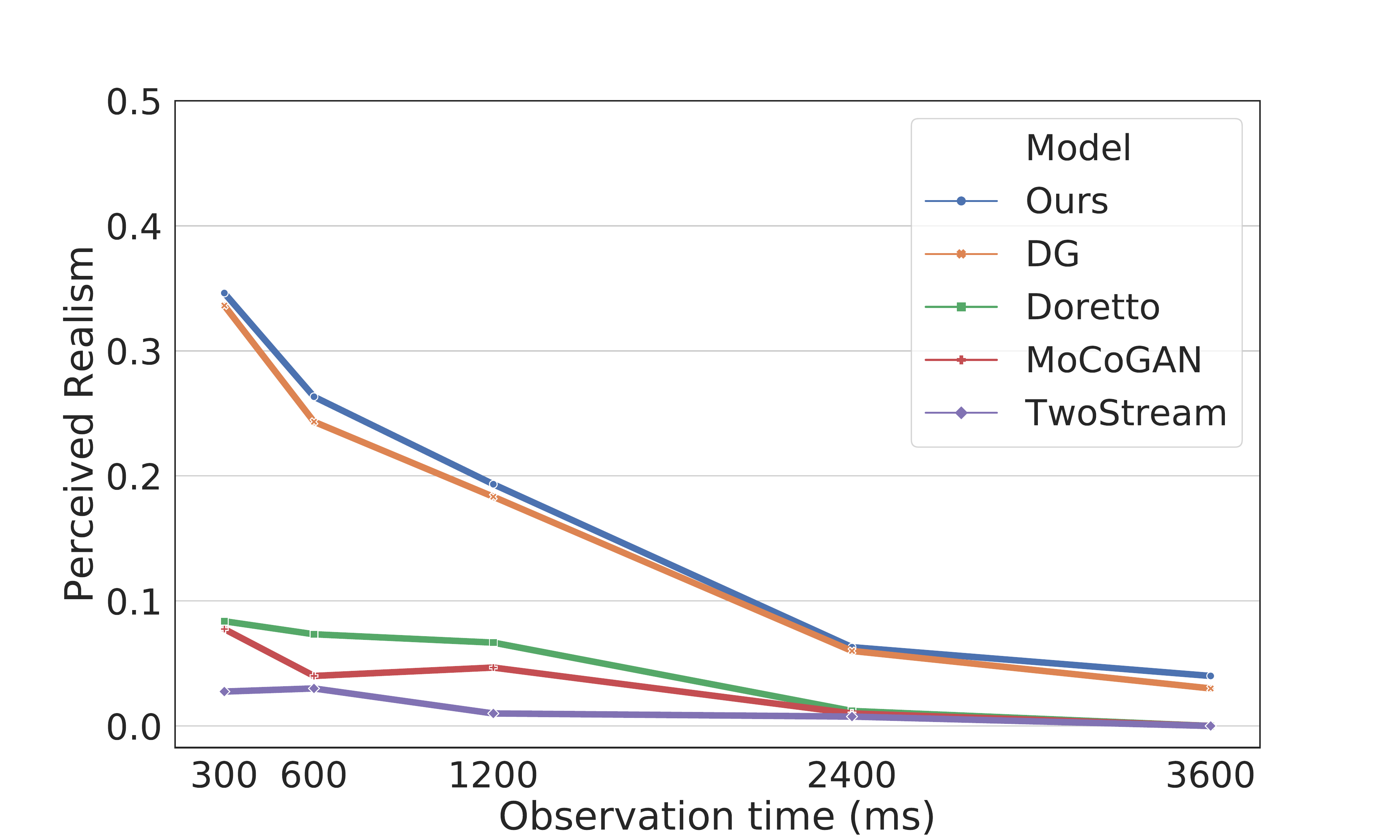}
	\caption{Limited time pairwise comparison results. Each curve shows the perceived realism over different observation times (ms). The number of pairwise comparisons is 36. The number of participants is 20. }
	\label{fig:human_study}
\end{center}
\end{figure}

\begin{figure}[t!]
\begin{center}
\hspace{0.5mm}\rotatebox{90}{\hspace{4mm}{\footnotesize obs1 }}
\includegraphics[width=.14\linewidth]{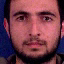}
\includegraphics[width=.14\linewidth]{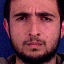}
\includegraphics[width=.14\linewidth]{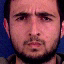}
\includegraphics[width=.14\linewidth]{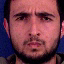}
\includegraphics[width=.14\linewidth]{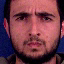}
\includegraphics[width=.14\linewidth]{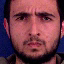}\\ \vspace{0.5mm}
\hspace{0.5mm}\rotatebox{90}{\hspace{4mm}{\footnotesize obs2 }}
\includegraphics[width=.14\linewidth]{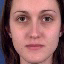}
\includegraphics[width=.14\linewidth]{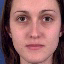}
\includegraphics[width=.14\linewidth]{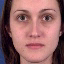}
\includegraphics[width=.14\linewidth]{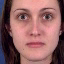}
\includegraphics[width=.14\linewidth]{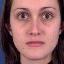}
\includegraphics[width=.14\linewidth]{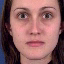}\\ 
(a) observed videos \\ \vspace{0.5mm}
\hspace{0.5mm}\rotatebox{90}{\hspace{1mm}{\footnotesize motion1 }}
\includegraphics[width=.14\linewidth]{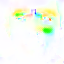}
\includegraphics[width=.14\linewidth]{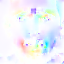}
\includegraphics[width=.14\linewidth]{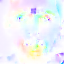}
\includegraphics[width=.14\linewidth]{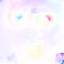}
\includegraphics[width=.14\linewidth]{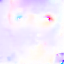}
\includegraphics[width=.14\linewidth]{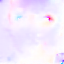}\\ \vspace{0.5mm}
\hspace{0.5mm}\rotatebox{90}{\hspace{1mm}{\footnotesize motion2 }}
\includegraphics[width=.14\linewidth]{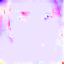}
\includegraphics[width=.14\linewidth]{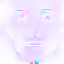}
\includegraphics[width=.14\linewidth]{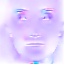}
\includegraphics[width=.14\linewidth]{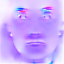}
\includegraphics[width=.14\linewidth]{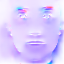}
\includegraphics[width=.14\linewidth]{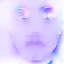}\\ \vspace{0.5mm}
(b) motions learned from training videos by our model\\ \vspace{1.5mm}
\hspace{0.5mm}\rotatebox{90}{\hspace{1mm}{\footnotesize o1 + m2 }}
\includegraphics[width=.14\linewidth]{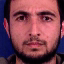}
\includegraphics[width=.14\linewidth]{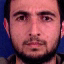}
\includegraphics[width=.14\linewidth]{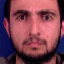}
\includegraphics[width=.14\linewidth]{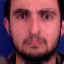}
\includegraphics[width=.14\linewidth]{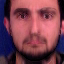}
\includegraphics[width=.14\linewidth]{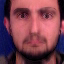}\\ \vspace{0.5mm}
\hspace{0.5mm}\rotatebox{90}{\hspace{1mm}{\footnotesize o2 + m1 }}
\includegraphics[width=.14\linewidth]{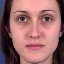}
\includegraphics[width=.14\linewidth]{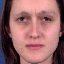}
\includegraphics[width=.14\linewidth]{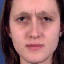}
\includegraphics[width=.14\linewidth]{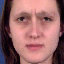}
\includegraphics[width=.14\linewidth]{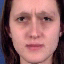}
\includegraphics[width=.14\linewidth]{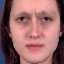}\\
(c) exchanging motions
\caption{Motion exchanging between two different facial appearances. (a) shows two observed facial expression videos. (b) displays the motions inferred from the observed videos by our model. (c) shows the results of exchanging the motions between two facial appearances.}
\label{fig:face_motion_exchanging}
\end{center}
\end{figure}

Some results of dynamic texture synthesis are displayed in Figure \ref{fig:synthesis}. We show the synthesis results by displaying the frames in the video sequences. For each example, the first row displays 6 frames of the observed 30-frame video sequence, while the second and the third rows show the corresponding 6 frames of two synthesized 30-frame video sequences that are generated by the learned model.   

Human perception has been used in \cite{chen2017photographic,tesfaldet2018two,wang2018high,XieGaoZhengZhuWu2019} to evaluate the synthesis quality. We follow the same protocol in \cite{XieGaoZhengZhuWu2019} to conduct a human perceptual study to get feedback from human subjects on evaluating the visual quality of the generated dynamic textures. We randomly choose 20 different human observers to participate in the perceptual test, where each participant needs to perform 36 (12 categories $\times$ 3 examples per category) pairwise comparisons between a synthesized dynamic texture and its real version. For each pairwise comparison, participants are asked to select the more realistic one after observing each pair of dynamic textures for a specified observation time, which is chosen from discrete durations between 0.3 and 3.6 seconds. The varying observation time will help us to investigate how quickly the difference between dynamic textures can be identified. We specifically ask the participants to carefully check for both temporal coherence and image quality. We present all the dynamic textures to the participants in the form of video with a resolution of $64 \times 64$ pixels. To obtain unbiased and reliable results, we randomize the comparisons across the left/right layout of two videos in each pair and the display order of different video pairs. We measure the realism of dynamic textures by the participant error rate in distinguishing synthesized dynamic textures from real ones. The higher the participant error rate, the more realistic the synthesized dynamic textures. The ``perfectly'' synthesized results would cause an error rate of 50$\%$, because random guesses are made when the participants are incapable of distinguishing the synthesized examples from the real ones.

For comparison, we use three baseline methods, such as LDS (linear dynamic system) \cite{doretto2003dynamic}, TwoStream \cite{tesfaldet2018two}, MoCoGAN \cite{tulyakov2018mocogan}, and dynamic generator (DG) \cite{XieGaoZhengZhuWu2019}. 
The comparison is performed on 12 dynamic texture videos (e.g., waterfall, burning fire, waving flag, etc) that have been used in \cite{XieGaoZhengZhuWu2019}.

\begin{figure}[t!]
\begin{center}
\hspace{0.5mm}\rotatebox{90}{\hspace{4mm}{\footnotesize obs }}
\includegraphics[width=.14\linewidth]{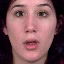}
\includegraphics[width=.14\linewidth]{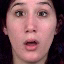}
\includegraphics[width=.14\linewidth]{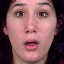}
\includegraphics[width=.14\linewidth]{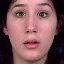}
\includegraphics[width=.14\linewidth]{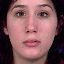}
\includegraphics[width=.14\linewidth]{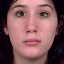}\\ 
{\footnotesize (a) observed facial expression (surprise)} \\
\vspace{1.5mm}
\hspace{0.5mm}\rotatebox{90}{\hspace{2mm}{\footnotesize motion }}
\includegraphics[width=.14\linewidth]{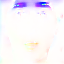}
\includegraphics[width=.14\linewidth]{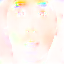}
\includegraphics[width=.14\linewidth]{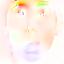}
\includegraphics[width=.14\linewidth]{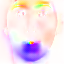}
\includegraphics[width=.14\linewidth]{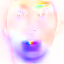}
\includegraphics[width=.14\linewidth]{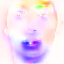}\\
{\footnotesize (b) learned motion (surprise)}\\
\vspace{1.5mm}
\hspace{0.5mm}\rotatebox{90}{\hspace{4mm}{\footnotesize syn1 }}
\includegraphics[width=.14\linewidth]{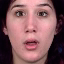}
\includegraphics[width=.14\linewidth]{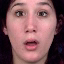}
\includegraphics[width=.14\linewidth]{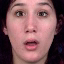}
\includegraphics[width=.14\linewidth]{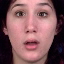}
\includegraphics[width=.14\linewidth]{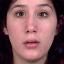}
\includegraphics[width=.14\linewidth]{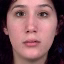}\\
\vspace{0.5mm}
\hspace{0.5mm}\rotatebox{90}{\hspace{4mm}{\footnotesize syn2 }}
\includegraphics[width=.14\linewidth]{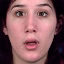}
\includegraphics[width=.14\linewidth]{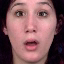}
\includegraphics[width=.14\linewidth]{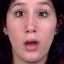}
\includegraphics[width=.14\linewidth]{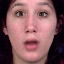}
\includegraphics[width=.14\linewidth]{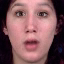}
\includegraphics[width=.14\linewidth]{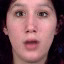}\\ 
{\footnotesize (c) synthesize 2 new ``surprise'' motions on the same appearance}  \\
\vspace{1.5mm}
\hspace{0.5mm}\rotatebox{90}{\hspace{4mm}{\footnotesize test1 }}
\includegraphics[width=.14\linewidth]{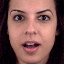}
\includegraphics[width=.14\linewidth]{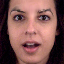}
\includegraphics[width=.14\linewidth]{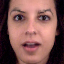}
\includegraphics[width=.14\linewidth]{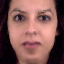}
\includegraphics[width=.14\linewidth]{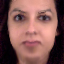}
\includegraphics[width=.14\linewidth]{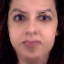}\\
\vspace{0.5mm}
\hspace{0.5mm}\rotatebox{90}{\hspace{4mm}{\footnotesize test2 }}
\includegraphics[width=.14\linewidth]{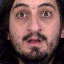}
\includegraphics[width=.14\linewidth]{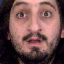}
\includegraphics[width=.14\linewidth]{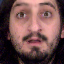}
\includegraphics[width=.14\linewidth]{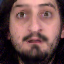}
\includegraphics[width=.14\linewidth]{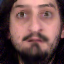}
\includegraphics[width=.14\linewidth]{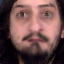}\\ 
{\footnotesize (d) apply the learned motion to some appearances in the testing set} \\
\vspace{1.5mm}
\hspace{0.5mm}\rotatebox{90}{\hspace{4mm}{\footnotesize test3 }}
\includegraphics[width=.14\linewidth]{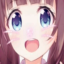}
\includegraphics[width=.14\linewidth]{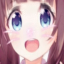}
\includegraphics[width=.14\linewidth]{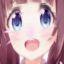}
\includegraphics[width=.14\linewidth]{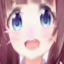}
\includegraphics[width=.14\linewidth]{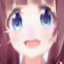}
\includegraphics[width=.14\linewidth]{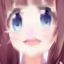}\\
\vspace{0.5mm}
\hspace{0.5mm}\rotatebox{90}{\hspace{4mm}{\footnotesize test4 }}
\includegraphics[width=.14\linewidth]{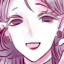}
\includegraphics[width=.14\linewidth]{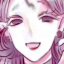}
\includegraphics[width=.14\linewidth]{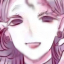}
\includegraphics[width=.14\linewidth]{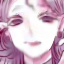}
\includegraphics[width=.14\linewidth]{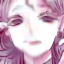}
\includegraphics[width=.14\linewidth]{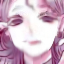} \\
{\footnotesize (e) apply the learned motion to cartoon appearances} \\
\vspace{1.5mm}
\hspace{0.5mm}\rotatebox{90}{\hspace{4mm}{\footnotesize test5 }}
\includegraphics[width=.14\linewidth]{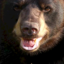}
\includegraphics[width=.14\linewidth]{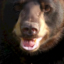}
\includegraphics[width=.14\linewidth]{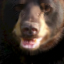}
\includegraphics[width=.14\linewidth]{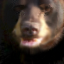}
\includegraphics[width=.14\linewidth]{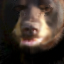}
\includegraphics[width=.14\linewidth]{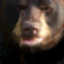} \\
\vspace{0.5mm} 
\hspace{0.5mm}\rotatebox{90}{\hspace{4mm}{\footnotesize test6 }}
\includegraphics[width=.14\linewidth]{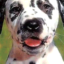}
\includegraphics[width=.14\linewidth]{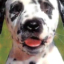}
\includegraphics[width=.14\linewidth]{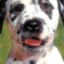}
\includegraphics[width=.14\linewidth]{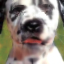}
\includegraphics[width=.14\linewidth]{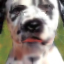}
\includegraphics[width=.14\linewidth]{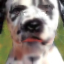} \\
{\footnotesize (f) apply the learned motion to animal appearances}
\vspace{1.5mm}
\caption{Transferring motion to new appearance. (a) shows some image frames of one observed facial motion video. (b) The learned motion from the observed video. (c) The synthesized new ``surprise'' motions on the same appearance. (d) The motion learned from the observed video is transferred to some new appearances extracted from videos in the testing set. (e) The learned motion is transferred to some cartoon appearances collected from Internet. (f) The learned motion is transferred to some animal faces collected from Internet.}
\label{fig:face_motion1}
\end{center}
\end{figure}

The results of this study are summarized in Figure \ref{fig:human_study}, which shows perceived realism (i.e., user error) as a function of observation time across methods.  Overall, the ``perceived realism'' decreases as observation time increases, and then stays at relatively the same level for longer observation. This means that as the observation time becomes longer, the participants feel easier to distinguish ``fake'' examples from real ones. The results clearly show that the dynamic textures generated by our models are more realistic than those obtained by models LDS, TwoStream, and MoCoGAN, and on par with those synthesized by DG.

To better understand the comparison results, we further analyze the performance of the baselines. We notice that the linear model (i.e., LDS) surpasses those methods using complicated deep network architecture (i.e., TwoStream and MoCoGAN). This is because one single training example is insufficient to train the MoCoGAN, which contains a large number of learning parameters, in an unstable adversarial learning scheme, while the TwoStream method, relying on pre-trained discriminative networks for feature matching, is incapable of synthesizing spatially inhomogeneous dynamic textures  (i.e., dynamic textures with structured background, e.g., boiling water in a static pot), which has been mentioned in \cite{tesfaldet2018two} and observed in \cite{XieGaoZhengZhuWu2019}. Our model is simple in the sense that it relies on neither auxiliary networks for variational or adversarial training nor pre-trained networks for feature matching, yet powerful in terms of disentanglement of appearance (represented by pixels), trackable motion (represented by pixel movements or optical flow), and intrackable motion (represented by residuals).

\begin{figure}[th!]
\begin{center}
\hspace{0.5mm}\rotatebox{90}{\hspace{4mm}{\footnotesize obs }}
\includegraphics[width=.14\linewidth]{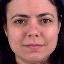}
\includegraphics[width=.14\linewidth]{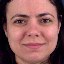}
\includegraphics[width=.14\linewidth]{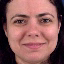}
\includegraphics[width=.14\linewidth]{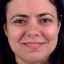}
\includegraphics[width=.14\linewidth]{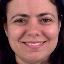}
\includegraphics[width=.14\linewidth]{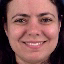}\\
{\footnotesize (a) observed facial expression (happiness)} \\
\hspace{0.5mm}\rotatebox{90}{\hspace{4mm}{\footnotesize motion }}
\vspace{1.5mm}
\includegraphics[width=.14\linewidth]{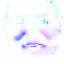}
\includegraphics[width=.14\linewidth]{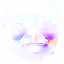}
\includegraphics[width=.14\linewidth]{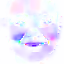}
\includegraphics[width=.14\linewidth]{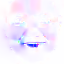}
\includegraphics[width=.14\linewidth]{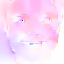}
\includegraphics[width=.14\linewidth]{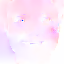}
\\
{\footnotesize (b) learned motion (happiness)} \\
\vspace{1.5mm}

\hspace{0.5mm}\rotatebox{90}{\hspace{4mm}{\footnotesize test1 }}
\includegraphics[width=.14\linewidth]{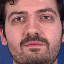}
\includegraphics[width=.14\linewidth]{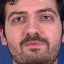}
\includegraphics[width=.14\linewidth]{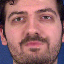}
\includegraphics[width=.14\linewidth]{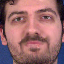}
\includegraphics[width=.14\linewidth]{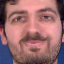}
\includegraphics[width=.14\linewidth]{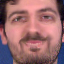}\\
\vspace{0.5mm}
\hspace{0.5mm}\rotatebox{90}{\hspace{4mm}{\footnotesize test2 }}
\includegraphics[width=.14\linewidth]{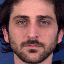}
\includegraphics[width=.14\linewidth]{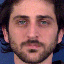}
\includegraphics[width=.14\linewidth]{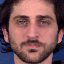}
\includegraphics[width=.14\linewidth]{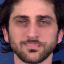}
\includegraphics[width=.14\linewidth]{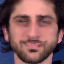}
\includegraphics[width=.14\linewidth]{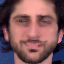}\\
{\footnotesize (c) apply the learned motion to some appearances in the testing set}  \\
\vspace{1.5mm}
\hspace{0.5mm}\rotatebox{90}{\hspace{4mm}{\footnotesize test3 }}
\includegraphics[width=.14\linewidth]{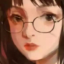}
\includegraphics[width=.14\linewidth]{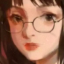}
\includegraphics[width=.14\linewidth]{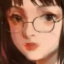}
\includegraphics[width=.14\linewidth]{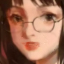}
\includegraphics[width=.14\linewidth]{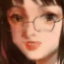}
\includegraphics[width=.14\linewidth]{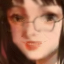}\\
\vspace{0.5mm}
\hspace{0.5mm}\rotatebox{90}{\hspace{4mm}{\footnotesize test4 }}
\includegraphics[width=.14\linewidth]{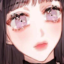}
\includegraphics[width=.14\linewidth]{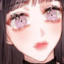}
\includegraphics[width=.14\linewidth]{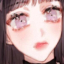}
\includegraphics[width=.14\linewidth]{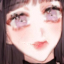}
\includegraphics[width=.14\linewidth]{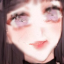}
\includegraphics[width=.14\linewidth]{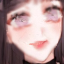}\\
{\footnotesize (d) apply the learned motion to cartoon appearances} \\
\vspace{1.5mm}
\hspace{0.5mm}\rotatebox{90}{\hspace{4mm}{\footnotesize test5 }}
\includegraphics[width=.14\linewidth]{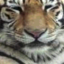}
\includegraphics[width=.14\linewidth]{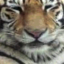}
\includegraphics[width=.14\linewidth]{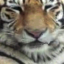}
\includegraphics[width=.14\linewidth]{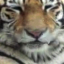}
\includegraphics[width=.14\linewidth]{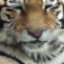}
\includegraphics[width=.14\linewidth]{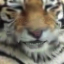} \\
\vspace{0.5mm} 
\hspace{0.5mm}\rotatebox{90}{\hspace{4mm}{\footnotesize test6 }}
\includegraphics[width=.14\linewidth]{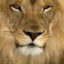}
\includegraphics[width=.14\linewidth]{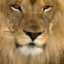}
\includegraphics[width=.14\linewidth]{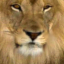}
\includegraphics[width=.14\linewidth]{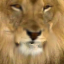}
\includegraphics[width=.14\linewidth]{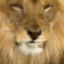}
\includegraphics[width=.14\linewidth]{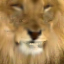} \\
{\footnotesize (e) apply the learned motion to animal appearances}
\vspace{1.5mm}
\caption{Transferring motion to new appearance. (a) shows some image frames of one observed facial motion video (happy). (b) The learned motion. (c) The motion learned from the observed video is transferred to some new appearances extracted from videos in the testing set. (d) The learned motion is transferred to some cartoon appearances collected from Internet. (e) The learned motion is transferred to some animal faces collected from Internet.}
\label{fig:face_motion2}
\end{center}
\end{figure}

\subsection{Experiment 2: Unsupervised disentanglement of appearance and motion}

To study the performance of the proposed model for disentanglement of appearance and motion, we perform a motion exchange experiment between two randomly selected facial expression sequences from MUG Facial Expression dataset \cite{MUG2010} by the learned model. We first  disentangle the appearance vector $c$, optical flow $\{M_t\}$ as trackable motion, and  residuals $\{R_t\}$ as intrackable motion for each of the two sequences by fitting our model on them. We then exchange their inffered motions $\{M_t\}$ and regenerate both sequences by repeatedly warping the appearance images that are generated by their own appearance vectors $c$ with the exchanged optical flows $\{M_t\}$.   
Figure \ref{fig:face_motion_exchanging} displays the results, where (a) shows some image frames of two selected original facial expression videos respectively. One is a man with sadness facial expression, and the other is a woman with surprise facial expression; (b) visualizes the learned trackable motions (optical flows) by  color images, by following \cite{baker2011database}, where each color represents a direction (Please see the appendix for the displacement field color coding map.); and (c) displays some image frames of the generated videos after motion exchange between the man and the woman.

From Figure \ref{fig:face_motion_exchanging}, we can see that the motion latent vectors do not encode any appearance information. The color, illumination, and identity information in the generated video sequence only depend on the appearance latent vector, and are not changed after motion exchange. Figure \ref{fig:face_motion1}  demonstrates an idea of learning from only one single video and unsupervisedly disentangling the motion and appearance of the video, and then transferring the motion to the other appearances.   Figure \ref{fig:face_motion1} (a) displays some image frames of one observed video where a woman is performing surprise expression. We first learn a model from the observed video. Figure \ref{fig:face_motion1} (b) visualizes the learned motion. We then fix the inferred appearance latent vector and synthesize new surprise facial expression by randomly sampling the latent  vectors of the learned model. Two new synthesized ``surprise'' expressions on the same women are shown in Figure \ref{fig:face_motion1} (c). We further study transferring the learned motion to some unseen appearances. We select two unseen faces from the testing set. We apply the learned motion (i.e., the learned warping sequence) to the first frame of each testing video, and generate new image sequences, as shown in Figure \ref{fig:face_motion1} (d). We can also apply the learned motion to some faces from other domains. Figure \ref{fig:face_motion1} (e) shows two examples of transferring the learned motion to the cartoon face images. In each example, the image frame shown in the first column is the input appearance, and the rest image frames are generated when we apply the learned warping sequence to the input appearance.  We can even apply the learned human facial expression motion to non-human appearances, such as animal faces (see Figure \ref{fig:face_motion1}(f)). Figure \ref{fig:face_motion2} shows one more example of motion transfer from another input video. 

Although the appearance domain in testing is significantly different from that in training, because our trackable motion does not encode any appearance information, the motion transfer will not modify the appearance information, which corroborates the disentangling power of the proposed model. Currently, our model does not consider face geometric deformation. We assume the face data we used in this experiment are well aligned. We can easily pre-align a testing face by morphing, when performing motion transfer to a non-aligned testing face, and then morph the new generated faces in each image frame back to its original shape. More rigorously, we can add one more generator that takes care of the shape geometric deformation of the appearance to deal with the alignment issue. The training of such a model will lead to an unsupervised disentanglement of appearance, geometry, and motion of video. We leave this as our future work.   

\begin{figure}[t!]
\begin{center}
\hspace{0.5mm}\rotatebox{90}{\hspace{4mm}{\footnotesize obs }}
\includegraphics[width=.14\linewidth]{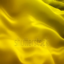}
\includegraphics[width=.14\linewidth]{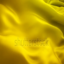}
\includegraphics[width=.14\linewidth]{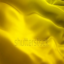}
\includegraphics[width=.14\linewidth]{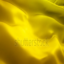}
\includegraphics[width=.14\linewidth]{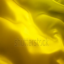}
\includegraphics[width=.14\linewidth]{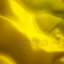}
 \\
(a) observed videos \\ \vspace{0.5mm}
\hspace{0.5mm}\rotatebox{90}{\hspace{2mm}{\footnotesize motion }}
\includegraphics[width=.14\linewidth]{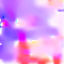}
\includegraphics[width=.14\linewidth]{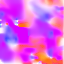}
\includegraphics[width=.14\linewidth]{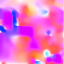}
\includegraphics[width=.14\linewidth]{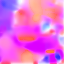}
\includegraphics[width=.14\linewidth]{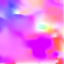}
\includegraphics[width=.14\linewidth]{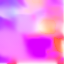}
\\
(b) learned motion (waving) \\ \vspace{0.5mm}
\hspace{0.5mm}\rotatebox{90}{\hspace{4mm}{\footnotesize test 1 }}
\includegraphics[width=.14\linewidth]{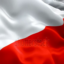}
\includegraphics[width=.14\linewidth]{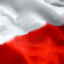}
\includegraphics[width=.14\linewidth]{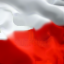}
\includegraphics[width=.14\linewidth]{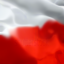}
\includegraphics[width=.14\linewidth]{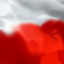}
\includegraphics[width=.14\linewidth]{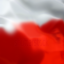} \\ \vspace{0.5mm}
\hspace{0.5mm}\rotatebox{90}{\hspace{4mm}{\footnotesize test 2 }}
\includegraphics[width=.14\linewidth]{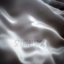}
\includegraphics[width=.14\linewidth]{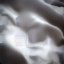}
\includegraphics[width=.14\linewidth]{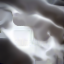}
\includegraphics[width=.14\linewidth]{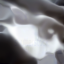}
\includegraphics[width=.14\linewidth]{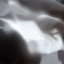}
\includegraphics[width=.14\linewidth]{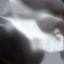} \\ 
(c) apply the learned motion to new appearances \\ \vspace{0.5mm}
\caption{Transferring new motion to new appearance (dynamic texture). (a) shows some image frames of one observed dynamic texture (waving flag). (b) The learned motion from the observed video. (c) The learned motion is transferred to some new flags. The given input appearance is shown in the first column of each example.}
\label{fig:flag_motion}
\end{center}
\end{figure}

Figure \ref{fig:flag_motion} shows another example of motion transfer from dynamic texture. Similarly, we learn our model from the waving yellow flag, which is shown in Figure \ref{fig:flag_motion}(a), and transfer the learned motion (shown in Figure \ref{fig:flag_motion}(b)) to some new images of flags to make them waving in Figure \ref{fig:flag_motion}(c). We can use the learned model to generate an arbitrarily long motion sequence and transfer it to different images. 

\subsection{Experiment 3: Unsupervised disentanglement of trackable and intrackable motions}

\begin{figure}[t!]
\begin{center}
\hspace{8mm} \includegraphics[width=.14\linewidth]{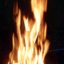}
\includegraphics[width=.14\linewidth]{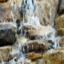}
\includegraphics[width=.14\linewidth]{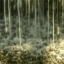}
\includegraphics[width=.14\linewidth]{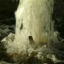}
\includegraphics[width=.14\linewidth]{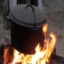}\\ \vspace{-1mm}
{\scriptsize \hspace{8mm}  (1) \hspace{9mm}(2) \hspace{8mm}(3) \hspace{9mm}(4) \hspace{9mm}(5)}\\ \vspace{1mm}
\hspace{9mm} \includegraphics[width=.14\linewidth]{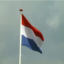}
\includegraphics[width=.14\linewidth]{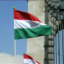}
\includegraphics[width=.14\linewidth]{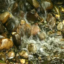}
\includegraphics[width=.14\linewidth]{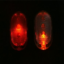}
\includegraphics[width=.14\linewidth]{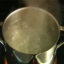}\\ \vspace{-1mm}
{\scriptsize \hspace{8mm}  (6) \hspace{9mm}(7) \hspace{9mm}(8) \hspace{9mm}(9) \hspace{9mm}(10)}\\ \vspace{1mm}
\includegraphics[width=.83\linewidth]{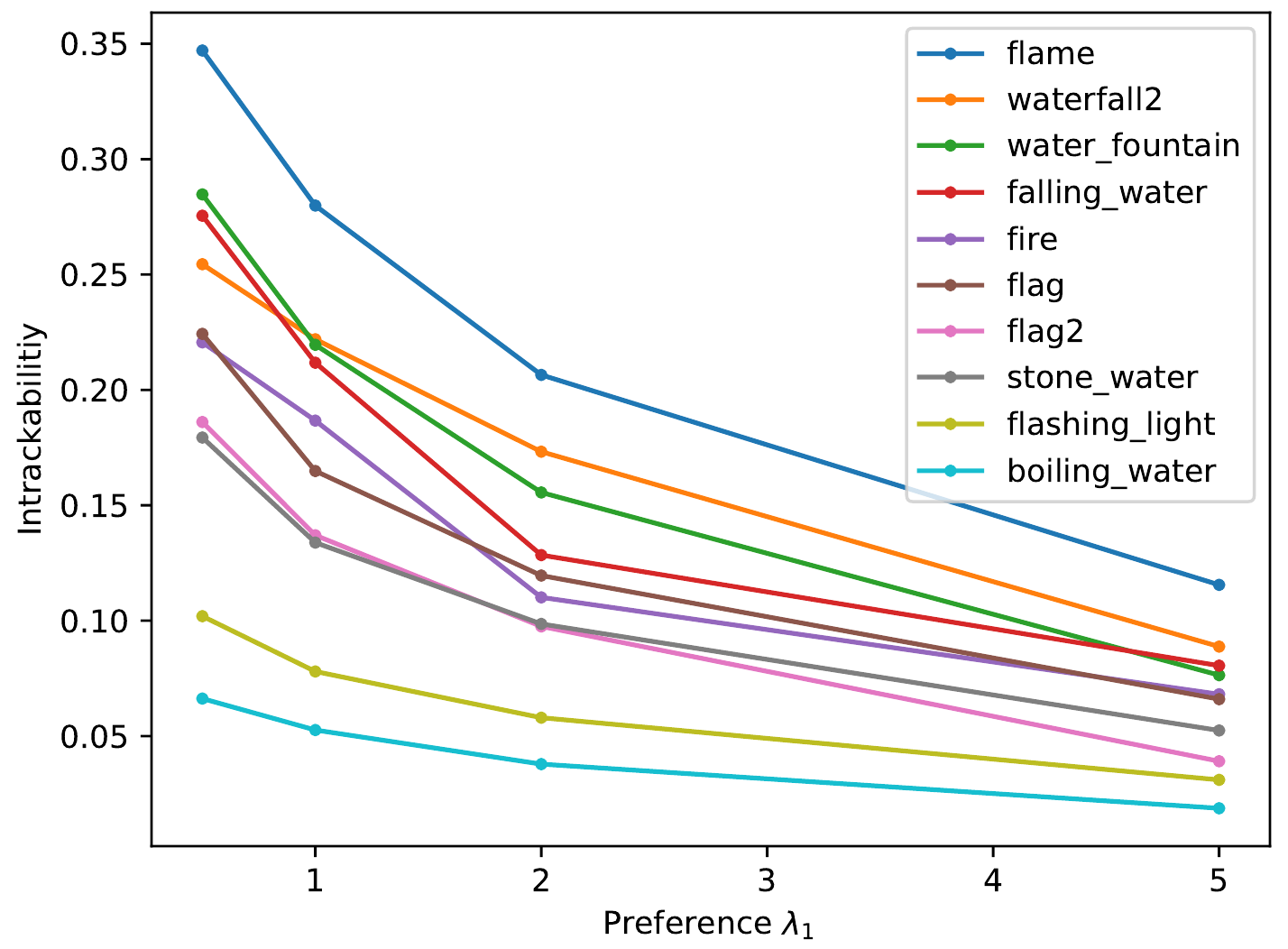}
\caption{The intrackability score vs the preference rate $\lambda_1$ (the penalty parameter for the norm of the residual image). The penalty parameter for smoothness is 0.005.}
\label{fig:trackability}
\end{center}
\end{figure}

Intrackability (or trackability) is an important concept of motion patterns, which has been studied in \cite{gong2012intrackability}. It was demonstrated in \cite{wu2004information,gong2012intrackability,han2015video} that trackability changes over scales, densities, and stochasticity of the dynamics. For example, trackability of a video of waterfall will depend on the distance between the observed target and the observer. Besides, the observer's preference for interpreting dynamic motions via tracking appearance details is a subjective factor to affect the perceived trackability of a dynamic pattern in the visual system of the brain. 

In the context of our model, we can define intractability as the ratio between the average of $\ell_2$ norm of the non-tractable residual image $R_t$ and the average of the $\ell_2$ norm of the observed image $\I_t$. This ratio depends on the penalty parameter $\lambda_1$ of the $\ell_2$ norm of $R_t$ used in the learning stage. This penalty parameter corresponds to the subjective preference
mentioned above. The larger the preference $\lambda_1$ is, the larger extent to which we interpret a video by trackable contents, the less the residuals, and the less intrackability score. 

Our model can unsupervisedly disentangle the trackable and intrackable components of the training videos. The intrackability can be directly obtained as a result of learning the model, where we do not need the ground truth or pre-inferred optical flows. In addition, the intrackability is defined in terms of the coherent motion pattern learned from the whole video sequence by our model.

\begin{figure}[t!]
\begin{center}
\hspace{1.5mm}\rotatebox{90}{\hspace{2mm}{\tiny original}}
\includegraphics[width=.14\linewidth]{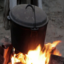}
\includegraphics[width=.14\linewidth]{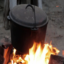}
\includegraphics[width=.14\linewidth]{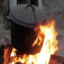}
\includegraphics[width=.14\linewidth]{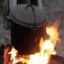}
\includegraphics[width=.14\linewidth]{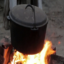}
\includegraphics[width=.14\linewidth]{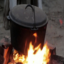}\\
(a) observed video\\
\vspace{1mm}
\hspace{1.5mm}\rotatebox{90}{\hspace{2mm}{\tiny trackable }}
\includegraphics[width=.14\linewidth]{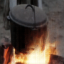}
\includegraphics[width=.14\linewidth]{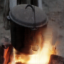}
\includegraphics[width=.14\linewidth]{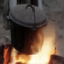}
\includegraphics[width=.14\linewidth]{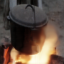}
\includegraphics[width=.14\linewidth]{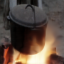}
\includegraphics[width=.14\linewidth]{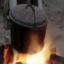}\\
\vspace{1mm}
\hspace{1.5mm}\rotatebox{90}{{\tiny optical flow}}
\includegraphics[width=.14\linewidth]{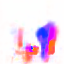}
\includegraphics[width=.14\linewidth]{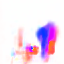}
\includegraphics[width=.14\linewidth]{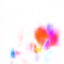}
\includegraphics[width=.14\linewidth]{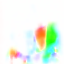}
\includegraphics[width=.14\linewidth]{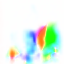}
\includegraphics[width=.14\linewidth]{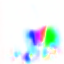}\\ \vspace{1mm}
\hspace{1.5mm}\rotatebox{90}{{\tiny intrackable}}
\includegraphics[width=.14\linewidth]{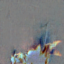}
\includegraphics[width=.14\linewidth]{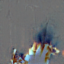}
\includegraphics[width=.14\linewidth]{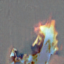}
\includegraphics[width=.14\linewidth]{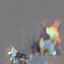}
\includegraphics[width=.14\linewidth]{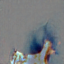}
\includegraphics[width=.14\linewidth]{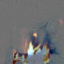}\\
(b) preference rate $\lambda_1 = 0.5$\\ \vspace{1mm}
\vspace{1mm}
\hspace{1.5mm}\rotatebox{90}{\hspace{2mm}{\tiny trackable }}
\includegraphics[width=.14\linewidth]{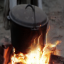}
\includegraphics[width=.14\linewidth]{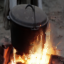}
\includegraphics[width=.14\linewidth]{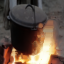}
\includegraphics[width=.14\linewidth]{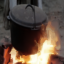}
\includegraphics[width=.14\linewidth]{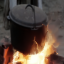}
\includegraphics[width=.14\linewidth]{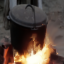}\\
\vspace{1mm}
\hspace{1.5mm}\rotatebox{90}{{\tiny optical flow}}
\includegraphics[width=.14\linewidth]{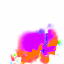}
\includegraphics[width=.14\linewidth]{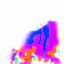}
\includegraphics[width=.14\linewidth]{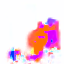}
\includegraphics[width=.14\linewidth]{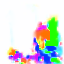}
\includegraphics[width=.14\linewidth]{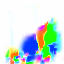}
\includegraphics[width=.14\linewidth]{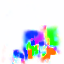}\\ \vspace{1mm}
\hspace{1.5mm}\rotatebox{90}{{\tiny intrackable}}
\includegraphics[width=.14\linewidth]{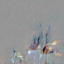}
\includegraphics[width=.14\linewidth]{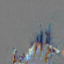}
\includegraphics[width=.14\linewidth]{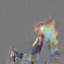}
\includegraphics[width=.14\linewidth]{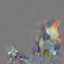}
\includegraphics[width=.14\linewidth]{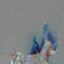}
\includegraphics[width=.14\linewidth]{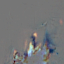}\\
(c) preference rate $\lambda_1 = 5$\\
\caption{Unsupervised disentanglement of trackable and intrackable motions of a video exhibiting a burning fire heating a pot. (a) displays the training video. In each of panels (b) and (c), the first row shows trackable component, the second row shows the corresponding optical flows, and the third row shows the corresponding intrackable component. The penalty parameter for smoothness is 0.005.}
\label{fig:tack_non}
\end{center}
\end{figure}

Figure \ref{fig:trackability} shows a curve of intrackability scores under different preference rates ($\lambda_1=0.5, 1, 2$ and $5$) for each of 10 different dynamic patterns. One typical image frame is illustrated for each of video clips that we used. The model structure and hyperparameter setting are the same as the one we used in Experiment 1. The penalty parameter for smoothness is fixed to be 0.005. The results are reasonable and consistent with our empirical observations and intuitions. For example, under the same subjective preference, a video with structured background and slow motion tends to have a lower intrackability score because one can track the elements in motion easily (e.g., a video clip exhibiting boiling water in a static pot), while a video with fast and random motion tends to have a higher intrackability score due to the loss of track of the elements in the video (e.g., a video clip exhibiting burning flame or flowing water). Moreover, we find that as the preference $\lambda_1$ increases, the intrackability of all videos decrease, because the model seeks to interpret each video using more trackable motion.

\begin{figure}[t!]
\begin{center}
\hspace{1.5mm}\rotatebox{90}{\hspace{2mm}{\tiny original}}
\includegraphics[width=.14\linewidth]{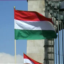}
\includegraphics[width=.14\linewidth]{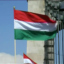}
\includegraphics[width=.14\linewidth]{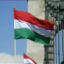}
\includegraphics[width=.14\linewidth]{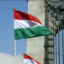}
\includegraphics[width=.14\linewidth]{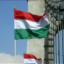}
\includegraphics[width=.14\linewidth]{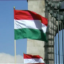}\\
(a) observed video\\
\vspace{1mm}
\hspace{1.5mm}\rotatebox{90}{\hspace{2mm}{\tiny trackable }}
\includegraphics[width=.14\linewidth]{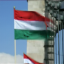}
\includegraphics[width=.14\linewidth]{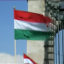}
\includegraphics[width=.14\linewidth]{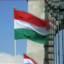}
\includegraphics[width=.14\linewidth]{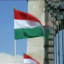}
\includegraphics[width=.14\linewidth]{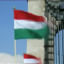}
\includegraphics[width=.14\linewidth]{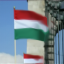}\\
\vspace{1mm}
\hspace{1.5mm}\rotatebox{90}{{\tiny optical flow}}
\includegraphics[width=.14\linewidth]{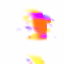}
\includegraphics[width=.14\linewidth]{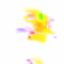}
\includegraphics[width=.14\linewidth]{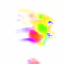}
\includegraphics[width=.14\linewidth]{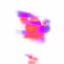}
\includegraphics[width=.14\linewidth]{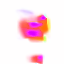}
\includegraphics[width=.14\linewidth]{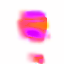}\\ \vspace{1mm}
\hspace{1.5mm}\rotatebox{90}{{\tiny intrackable}}
\includegraphics[width=.14\linewidth]{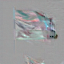}
\includegraphics[width=.14\linewidth]{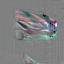}
\includegraphics[width=.14\linewidth]{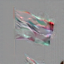}
\includegraphics[width=.14\linewidth]{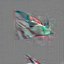}
\includegraphics[width=.14\linewidth]{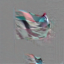}
\includegraphics[width=.14\linewidth]{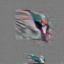}\\
(b) preference rate $\lambda_1 = 0.5$\\ \vspace{1mm}
\vspace{1mm}
\hspace{1.5mm}\rotatebox{90}{\hspace{2mm}{\tiny trackable }}
\includegraphics[width=.14\linewidth]{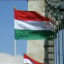}
\includegraphics[width=.14\linewidth]{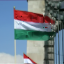}
\includegraphics[width=.14\linewidth]{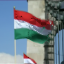}
\includegraphics[width=.14\linewidth]{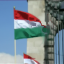}
\includegraphics[width=.14\linewidth]{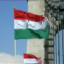}
\includegraphics[width=.14\linewidth]{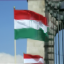}\\
\vspace{1mm}
\hspace{1.5mm}\rotatebox{90}{{\tiny optical flow}}
\includegraphics[width=.14\linewidth]{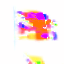}
\includegraphics[width=.14\linewidth]{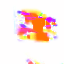}
\includegraphics[width=.14\linewidth]{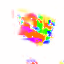}
\includegraphics[width=.14\linewidth]{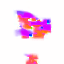}
\includegraphics[width=.14\linewidth]{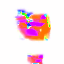}
\includegraphics[width=.14\linewidth]{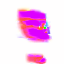}\\ \vspace{1mm}
\hspace{1.5mm}\rotatebox{90}{{\tiny intrackable}}
\includegraphics[width=.14\linewidth]{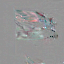}
\includegraphics[width=.14\linewidth]{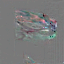}
\includegraphics[width=.14\linewidth]{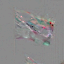}
\includegraphics[width=.14\linewidth]{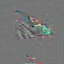}
\includegraphics[width=.14\linewidth]{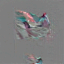}
\includegraphics[width=.14\linewidth]{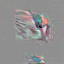}
\\
(c) preference rate $\lambda_1 = 5$\\
\caption{Unsupervised disentanglement of trackable and intrackable motions of a waving flag video. (a) displays the training video. In each of panels (b) and (c), the first row shows trackable component, the second row shows the corresponding optical flows, and the third row shows the corresponding intrackable component. The penalty parameter for smoothness is 0.005.}
\label{fig:tack_non2}
\end{center}
\end{figure}

Figure \ref{fig:tack_non} and \ref{fig:tack_non2} demonstrate two examples of unsupervised disentanglement of trackable and intrackable components from an observed video under different preference rates. In each of the figures, panel (a) displays some image frames of the training video, while panels (b) and (c) show the disentanglement results under preference rates equal to 0.5 and 5, respectively. We can see that the residual part (i.e., intrackable component) decreases and the optical flows (or displacement fields) become detailed and complicated, as the preference rate increases.  Our model is natural to understand the concept of intrackability of dynamic patterns.

We also conduct an ablation study to investigate the effect of the part of intrackable motion in our model, by comparing the full model with the one only taking into account the trackable motion. Table \ref{tab:ablation} reports the average training loss across 12 training videos with different training epochs. The results suggest that, with the same numbers of training epochs, the model without considering intrackable motion tends to have higher training loss, especially when the intrackability of the video is high. Thus, intrackable motion is  indispensable in representing a dynamic pattern.  

\begin{table}[t!]
\centering
\caption{An ablation study of the effect of intrackable motion. Training errors under different numbers of epochs are averaged across 12 training videos.}
\begin{tabular}{|c|c|c|c|c|}
\hline
epoch          & 2000            & 3000            & 4000            & 5000            \\ \hline
full model          & \textbf{0.0285} & \textbf{0.0253} & \textbf{0.0235} & \textbf{0.0222} \\ \hline
trackable only & 0.0487          & 0.0463         & 0.0442          & 0.0426          \\ \hline
\end{tabular}
\label{tab:ablation}
\end{table}

\section{Conclusion}

This paper proposes a motion-based generator model for dynamic patterns. The model is capable of disentangling the image sequence into appearance, trackable and intrackable motions, by modeling them by non-linear state space models, where the non-linear functions in the transition model and the emission model are parametrized by neural networks. 

A key feature of our model is that we can learn the model without ground truth or pre-inference of the movements of the pixels or the optical flows. They are automatically inferred in the learning process. We show that the learned model for the motion can be generalized to unseen images by animating them according to the learned motion pattern. We also show that in the context of the learned model, we can define the notion of intrackability of the training dynamic patterns. 

\section*{Project page} 
{\small The code and videos of our generated results can be found at 
\url{http://www.stat.ucla.edu/~jxie/MotionBasedGenerator/MotionBasedGenerator.html}}

\section*{Acknowledgement}
{\small The work is supported by DARPA XAI project N66001-17-2-4029; ARO project W911NF1810296; ONR MURI project N00014-16-1-2007. We thank Yifei Xu for his
assistance with experiments. We gratefully acknowledge the support of NVIDIA Corporation with the donation of the Titan Xp GPU used for this research.}

\section*{Appendix} 
\begin{figure}[h!]
\begin{center}
\includegraphics[width=.26\linewidth]{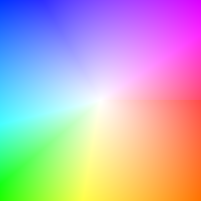} \hspace{2mm}
\includegraphics[width=.26\linewidth]{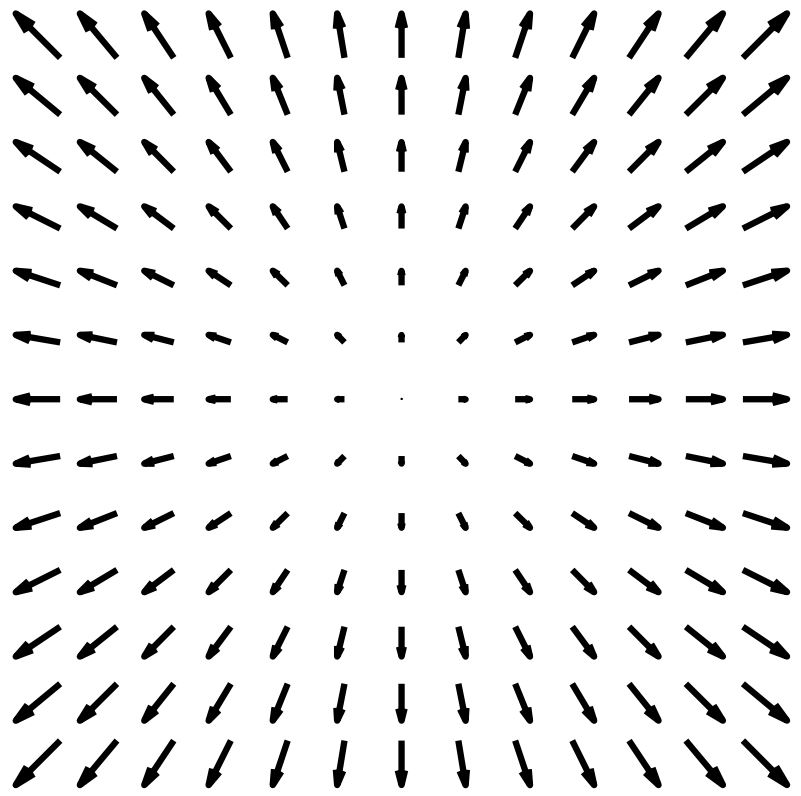}
\caption{Visualization of displacement field. The displacement of every pixel in this illustration is the vector from the center of the square to this pixel. }\label{fig:colormap1}
\end{center}
\end{figure}

Figure \ref{fig:colormap1} shows the color map for the color coded displacement fields used in \cite{liu2010sift}. We visualize  trackable motion (optical flow) by using the same color map in this paper.

\bibliographystyle{aaai}
{\footnotesize
\bibliography{mybibfile}}

\end{document}